\documentclass[letterpaper, 10 pt, conference]{ieeeconf}  

\IEEEoverridecommandlockouts                              
\usepackage[utf8]{inputenc}
\usepackage[T1]{fontenc}
\usepackage{epsfig} 
\usepackage{amsmath} 
\usepackage{amssymb}  
\usepackage{color}
\usepackage{cite}
\usepackage{caption}
\usepackage{aecompl}
\usepackage[T1]{fontenc}
\usepackage[normalem]{ulem} 

\usepackage{booktabs}
\usepackage{arydshln}
\usepackage{multirow}
\usepackage{mathrsfs}
\usepackage[dvipsnames]{xcolor}

\usepackage{soul}
\usepackage{ulem}
\newcounter{RNum}
\renewcommand{\theRNum}{\arabic{RNum}}
\newcommand{\Remark}{\noindent\textit{\textbf{Remark}~\refstepcounter{RNum}\textbf{\theRNum}: }}
\newcommand{\NoOne}[1]{\textcolor{red}{#1}}
\newcommand{\NoTwo}[1]{\textcolor{green}{#1}}
\newcommand{\NoThree}[1]{\textcolor{blue}{#1}}
\usepackage{xcolor}
\soulregister\NoOne7
\soulregister\NoTwo7
\soulregister\NoThree7
\soulregister\Remark7

\soulregister{\cite}7 
\soulregister{\citep}7 
\soulregister{\citet}7 
\soulregister{\ref}7
\soulregister{\pageref}7 

\usepackage[absolute,overlay]{textpos}
\usepackage{multicol}
\usepackage{graphicx}
\usepackage{amsmath}
\usepackage{amssymb}
\usepackage{xcolor,colortbl}
\usepackage{makecell}
\usepackage{pifont}
\usepackage{cancel}
\usepackage{mathtools}
\usepackage{adjustbox}

\newcommand{\eg}{\textit{e.g.}}

\definecolor{navy}{RGB}{0, 0, 128} 
\definecolor{maroon}{RGB}{128, 0, 0} 
\usepackage{newfloat}
\usepackage{listings}
\usepackage{algorithm}
\usepackage{algorithmic}
\usepackage{subcaption} 
\usepackage{float}
\usepackage{tabularx}

\definecolor{rblue}{rgb}{0,0.5,1}
\definecolor{awesome}{rgb}{1.0, 0.13, 0.32}
\definecolor{hollywoodcerise}{rgb}{0.96, 0.0, 0.63}
\definecolor{lasallegreen}{rgb}{0.03, 0.47, 0.19}
\definecolor{hanpurple}{rgb}{0.32, 0.09, 0.98}
\definecolor{green(pigment)}{rgb}{0.0, 0.65, 0.31}

\definecolor{mygray}{gray}{.9}
\usepackage{booktabs}

\makeatletter
\let\NAT@parse\undefined
\makeatother
\usepackage[pagebackref=false,breaklinks=true,colorlinks,bookmarks=false]{hyperref}
\hypersetup{colorlinks=true,linkcolor={red},citecolor={hanpurple},urlcolor={magenta}}

\title{\LARGE \bf
Out-of-Distribution Semantic Occupancy Prediction
}

\author{Yuheng Zhang$^{1,*}$, Mengfei Duan$^{1,*}$, Kunyu Peng$^{2,3,\dag}$, Yuhang Wang$^{1}$, Ruiping Liu$^{2}$, Fei Teng$^{1}$, Kai Luo$^{1}$,\\Zhiyong Li$^{1}$, and Kailun Yang$^{1,4,\dag}$%
\thanks{This work was supported in part by the National Natural Science Foundation of China (Grant No. 62473139), in part by the Hunan Provincial Research and Development Project (Grant No. 2025QK3019), and in part by the State Key Laboratory of Autonomous Intelligent Unmanned Systems (the opening project number ZZKF2025-2-10).}%
\thanks{$^{1}$The authors are with the School of Artificial Intelligence and Robotics, Hunan University, China (email: kailun.yang@hnu.edu.cn).}%
\thanks{$^{2}$The authors are also with the Institute for Anthropomatics and Robotics, Karlsruhe Institute of Technology, Germany (email: kunyu.peng@kit.edu).}
\thanks{$^{3}$The author is also with INSAIT, Sofia University ``St. Kliment Ohridski'', Bulgaria.}
\thanks{$^{4}$The author is also with the National Engineering Research Center of Robot Visual Perception and Control Technology, Hunan University, China.}
\thanks{$^{*}$Equal contribution.}
\thanks{$^{\dag}$Corresponding authors: Kailun Yang and Kunyu Peng.}
}

\begin{document}

\maketitle
\thispagestyle{empty}
\pagestyle{empty}

\begin{abstract}
3D semantic occupancy prediction is crucial for autonomous driving, providing a dense, semantically rich environmental representation. However, existing methods focus on in-distribution scenes, making them susceptible to Out-of-Distribution (OoD) objects and long-tail distributions, which increase the risk of undetected anomalies and misinterpretations, posing safety hazards. To address these challenges, we introduce the task of Out-of-Distribution Semantic Occupancy Prediction, targeting OoD detection in 3D voxel space. To fill dataset gaps, we propose Realistic Anomaly Augmentation that injects synthetic anomalies while preserving realistic spatial and occlusion patterns, enabling the creation of two datasets: VAA-KITTI and VAA-KITTI-360. We then propose OccOoD, a novel framework that integrates OoD detection into 3D semantic occupancy prediction, which uses Cross-Space Semantic Refinement (CSSR) to refine semantic predictions from complementary voxel and BEV representations, improving OoD detection. Experimental results demonstrate that OccOoD achieves an AuROC of 65.50\% and an AuPRC$_r$ of 31.83\% within a 1.2\,m radius, while maintaining competitive semantic occupancy prediction accuracy, significantly improving detection sensitivity for unknown obstacles, and validating strong generalization in real-world urban driving scenes. The established datasets and source code will be made publicly available at \url{https://github.com/7uHeng/OccOoD}.
\end{abstract}

\IEEEpeerreviewmaketitle
\section{Introduction}
\label{sec:intro}
3D scene understanding is fundamental to autonomous driving and intelligent transportation systems, crucial for motion planning and obstacle avoidance~\cite{hu2023planning}. 
The cost-effectiveness and ease of deployment of cameras have made vision-based 3D understanding methods a focal point of research~\cite{cao2022monoscene,zhang2023occformer}. 
Among these methods, 3D semantic occupancy prediction networks~\cite{li2023voxformer} are distinguished by their ability to accurately reconstruct geometry and semantic details. 
From a human-centric systems perspective, detecting unexpected hazards is essential for safe human-robot interaction in autonomous driving.

\begin{figure}[!t]
    \centering
    \includegraphics[width=0.8\linewidth]{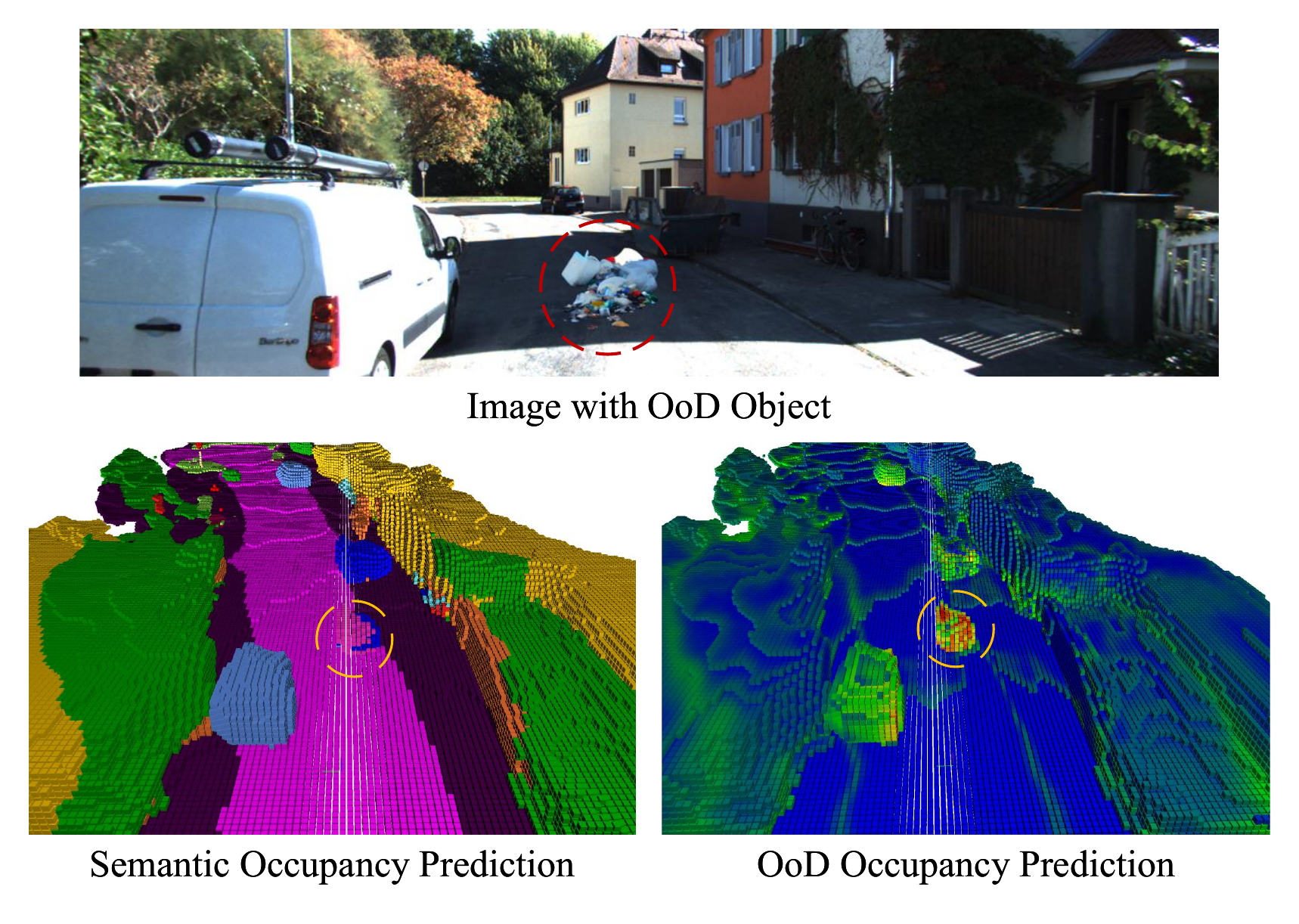}
    \vskip -2ex
    \captionof{figure} {{Visualization of the established task of \textbf{Out-of-Distribution Semantic Occupancy Prediction}. Due to the inherent limitation of closed-set prediction, mainstream semantic occupancy methods tend to force OoD objects into known categories, posing safety risks to autonomous driving. Proposed OccOoD accurately identifies OoD objects, with anomaly scores visualized from low (\textcolor{blue}{blue}) to high (\textcolor{red}{red}).
    }}
    \vskip -3.5ex
    \label{fig:teaser}
\end{figure}

However, current methods~\cite{mei2024camera,yu2024context} are constrained by closed-set training and evaluation, operating solely on predefined, fixed category sets. 
While effective in controlled settings, this approach~\cite{li2022bevformer} exposes systems to vulnerabilities from OoD objects and anomalies commonly encountered in real-world scenarios~\cite{blum2021fishyscapes,chan2021segmentmeifyoucan}, as seen in Fig.~\ref{fig:teaser}, where the standard semantic occupancy prediction method~\cite{mei2024camera} produces misclassifications.
OoD detection, the ability to identify entities that substantially deviate from the training data distribution, remains a critical yet underexplored challenge.
Such anomalies include unexpected obstacles (\eg, \emph{discarded furniture, stray wildlife, construction equipment}), unusual road conditions (\eg, \emph{potholes, debris, or oil spills}), and atypical weather (\eg, \emph{sudden fog, heavy rain, or snowdrifts}). As existing systems may misinterpret or fail to detect these deviations, they pose significant safety risks, potentially leading to failures and even catastrophic consequences in autonomous driving.

To bridge this critical gap, we introduce the task of Out-of-Distribution Semantic Occupancy Prediction, aiming to advance semantic scene understanding by integrating 3D occupancy prediction with OoD detection.
However, the lack of datasets annotated for both semantic occupancy and OoD objects hinders robust model development.
To address this, we propose a \textit{Realistic Anomaly Augmentation} that injects diverse anomalies into existing datasets, preserving realistic spatial and occlusion patterns. Crucially, the generated datasets are same-domain and designed to minimize domain shift with respect to the original data, requiring no additional supervision. 
This pipeline enables the creation of two novel datasets, VAA-KITTI and VAA-KITTI-360, comprising $26$ distinct categories of anomalies, where anomaly annotations encompass a diverse range of unforeseen hazards, facilitating comprehensive evaluation of OoD detection in 3D scene understanding. To explore real-world generalization, we further contribute VAA-STU, a real-world anomaly dataset for evaluation purposes.

Building on these datasets, we introduce OccOoD, the first unified framework that jointly performs semantic occupancy prediction and OoD detection.
First, we propose Cross-Space Semantic Refinement (CSSR), which refines semantic predictions by integrating complementary voxel and BEV features, where BEV provides enhanced global context to supplement local voxel representations, thereby improving both occupancy prediction robustness and OoD detection effectiveness.
Second, we introduce an OoD detection mechanism leveraging entropy and cosine similarity, enabling effective identification of anomalies in dynamic, unstructured scenes.
Furthermore, to ensure a comprehensive and realistic evaluation, we conduct extensive OoD detection experiments on the newly proposed datasets and validate on the real-world VAA-STU dataset, while evaluating occupancy prediction on public benchmarks. 
Experimental results show that OccOoD achieves AuROC scores of $65.50\%$ and $47.90\%$ on the proposed datasets, $65.37\%$ on VAA-STU, and mIoU scores of $16.80\%$ and $18.37\%$ on SemanticKITTI~\cite{behley2019semantickitti} and SSCBench-KITTI-360~\cite{li2024sscbench}, respectively. 
These results collectively establish OccOoD as a robust solution for addressing OoD challenges in 3D semantic occupancy prediction, demonstrating strong generalization and contributing to safer human-centric autonomous systems.

In a nutshell, our contributions are threefold:
\begin{itemize}
\item We first introduce OoD detection into semantic occupancy prediction, and propose 2D-prior Realistic Anomaly Synthesis to construct VAA-KITTI and VAA-KITTI-360 for autonomous driving safety evaluation
\item We propose OccOoD, a unified framework with CSSR that preserves voxel geometry and BEV context, enabling dual-space synergy for robust OoD discrimination without sacrificing occupancy accuracy.
\item Our proposed framework, OccOoD, achieves superior OoD detection performance on the established datasets while maintaining competitive accuracy in semantic occupancy prediction.
\end{itemize}

\section{Related Work}
\label{sec:formatting}

\noindent\textbf{Semantic Occupancy Prediction.}
3D semantic occupancy prediction
aims to infer both the geometry and semantics of a scene by inferring the occupancy state of voxels~\cite{song2017semantic}. 
To advance estimation quality and efficiency, efforts are made in 3D representation~\cite{liu2024fully_sparse,lu2025octreeocc}, 
context enhancement~\cite{wang2024h2gformer,li2024viewformer}, and long-sequence and generative modeling~\cite{wang2024occrwkv,li2024occmamba}. 
These efforts have collectively improved the quality and efficiency of semantic occupancy prediction.
Recently, several works~\cite{jiang2024symphonize, yu2024context} have introduced more fine-grained feature processing techniques. 
Meanwhile, uncertainty-aware methods~\cite{kalble2025evocc,heidrich2025occuq} incorporate uncertainty quantification to enhance estimation reliability.
However, these methods are limited to predefined categories, requiring retraining to generalize to novel object types and lacking flexibility.
While open-vocabulary learning~\cite{tan2023ovo, boeder2024langocc, vobecky2024pop_3d, zhang2025clip_occ} addresses generalization to arbitrary categories, it remains constrained to In-Distribution (ID) scenarios and fails to handle OoD detection. 
In real-world applications, the categories of anomalous objects are unforeseeable, making such methods less reliable. 
In contrast, our proposed model extends semantic occupancy prediction by perceiving all OoD objects, enabling efficient and comprehensive scene understanding.

\begin{figure*}
    \centering
    \includegraphics[width=0.85\linewidth]{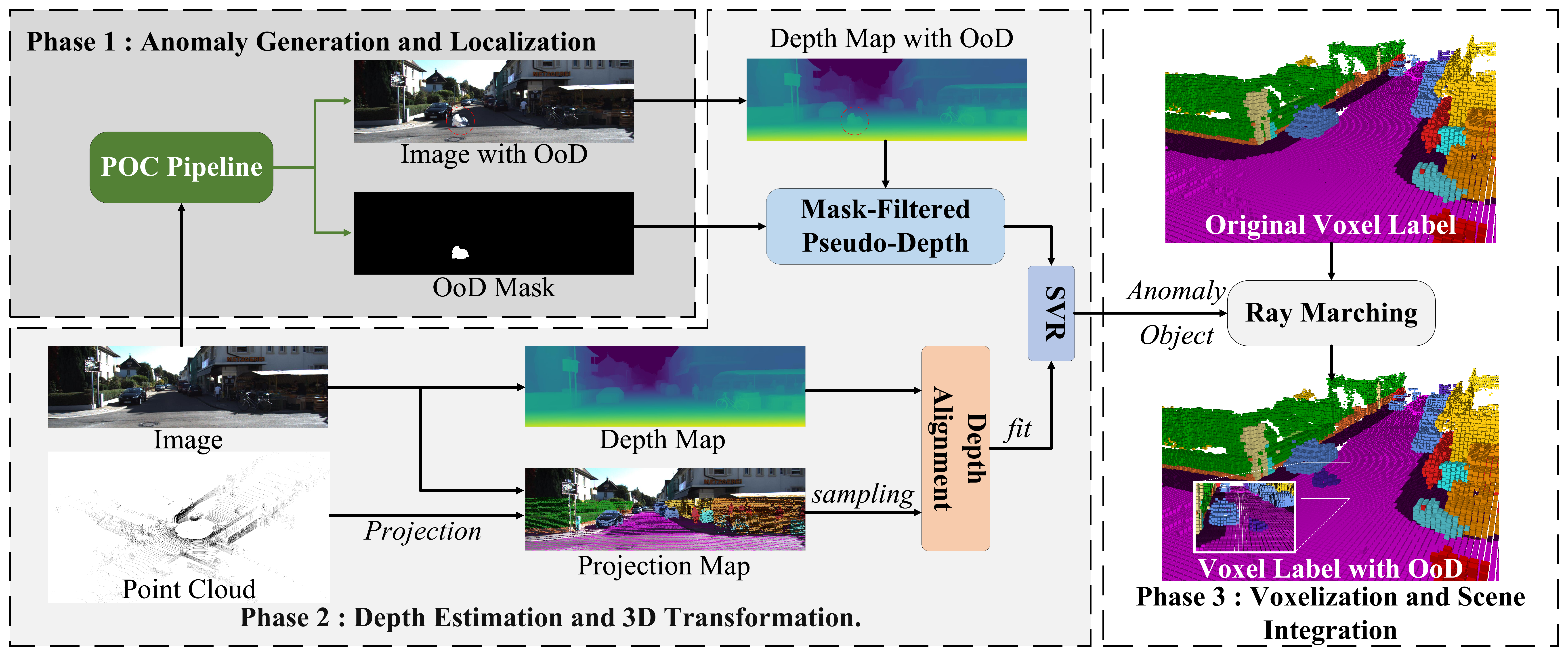}
    \vskip -1ex
    \caption{An illustration of the proposed Realistic Anomaly Augmentation. This pipeline is designed to address the challenges of collecting real-world OoD data by synthesizing anomalies that are physically plausible and contextually realistic.}
    \label{fig:Physics-Guided Anomaly Synthesis Pipeline}
    \vskip -3ex
\end{figure*}

\noindent\textbf{Out-of-Distribution Segmentation.} 
Out-of-distribution segmentation identifies anomalies in specific scenes~\cite{chan2021segmentmeifyoucan,blum2021fishyscapes} while maintaining in-distribution segmentation capabilities. Current research primarily focuses on anomaly detection in 2D images of road scenes, with the main methods including per-pixel architectures and mask-transformer-based approaches~\cite{rai2024mask2anomaly,nayal2023rba}. 
The per-pixel architecture methods include uncertainty-based~\cite{corbiere2019addressing,jung2021standardized}, reconstruction-based~\cite{lis2019detecting,grcic2021dense}, and outlier exposure~\cite{tian2022pixel,grcic2022densehybrid} methods.  
While advancing 2D detection, these methods are constrained by pinhole camera inputs, which limits their applicability in 3D scenarios demanding precise spatial perception. Some studies extend anomaly detection to 3D space, yet primarily target industrial~\cite{bergmann2021mvtec,liu2023real3d} or indoor object inspection~\cite{bhunia2024looking,li2024towards}, typically relying on static point clouds for defect localization~\cite{bergmann2023anomaly,zavrtanik2024cheating}. 
Due to their dependence on static geometric consistency and limited responsiveness to dynamic changes, these approaches are unsuitable for unconstrained, evolving environments. In contrast, we integrate out-of-distribution segmentation into vision-based 3D semantic occupancy prediction, enabling effective response to real-time changes and uncertainties, thereby overcoming the limitations of current methods in dynamic settings.

\section{Out-of-Distribution Datasets}
To address the lack of evaluation platforms, we propose a \textit{Realistic Anomaly Augmentation} to generate abnormal objects in real road scene images, yielding two same-domain datasets, \textbf{VAA-KITTI} and \textbf{VAA-KITTI-360}, which reflect semantic shifts rather than domain gaps for fair OoD evaluation. Alongside these, we introduce \textbf{VAA-STU}, derived from a real-world 3D point cloud dataset~\cite{nekrasov2025stu} containing natural outliers, with scenes voxelized for occupancy prediction, enabling evaluation under real-world OoD conditions.

\subsection{Realistic Anomaly Augmentation}

\begin{figure}[!t]
    \centering
    \includegraphics[width=0.95\linewidth]{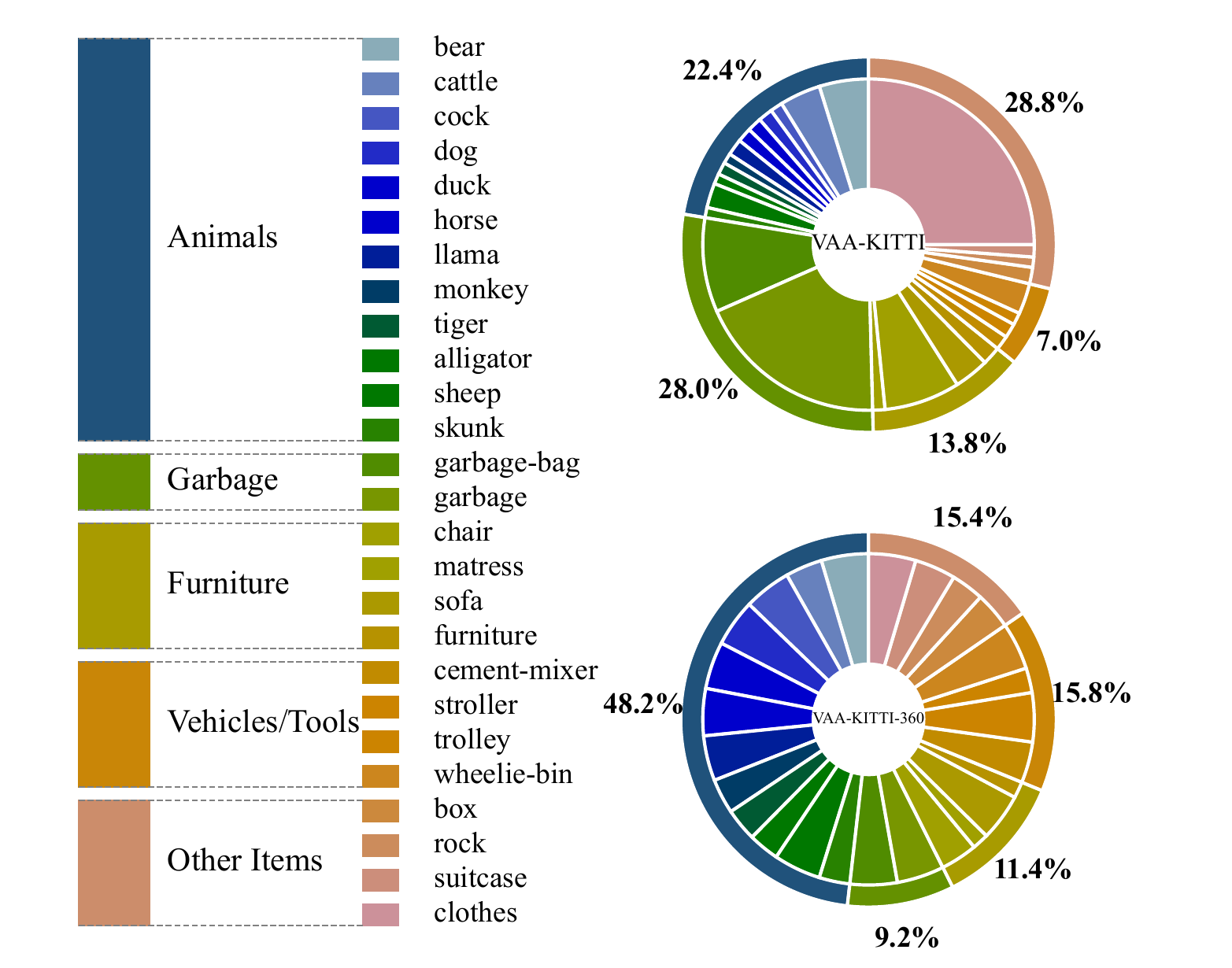}
    \vskip -2ex
    \caption{Distribution map of VAA-KITTI and VAA-KITTI-360, containing $26$ distinct OoD categories grouped into five main types as shown.}
    \label{fig:dataset_distribution}
    \vskip -3ex
\end{figure}

Collecting images and point cloud data of OoD objects is challenging due to their rarity in real-world scenes and the high cost of data collection and annotation. Moreover, 3D assets are expensive to acquire, and projecting them back to 2D for rendering often falls short of the fidelity achieved by native 2D generative methods. Meanwhile, simple Cut\&Paste techniques lack the realism offered by generative models. To address these limitations, we propose a \textit{Realistic Anomaly Augmentation} pipeline that generates synthetic anomalies under physical and environmental constraints, ensuring both plausibility and difficulty for robust OoD detection model evaluation. Our augmentation operates in the 2D image space to alleviate 3D data scarcity, leveraging the richness of 2D synthesis techniques while maintaining 3D geometric consistency. As illustrated in Fig.~\ref{fig:Physics-Guided Anomaly Synthesis Pipeline}, our pipeline is structured into three key phases:

\noindent\textbf{Phase 1: Anomaly Generation and Localization.} 
In 2D anomaly detection, studies have demonstrated that synthetic anomaly data is highly effective for model training and evaluation~\cite{de2024placing,duan2025panoramic}.
We employ POC~\cite{de2024placing}, a state-of-the-art image generation technique, to synthesize anomaly images and their corresponding mask maps. 
To better align with real-world application scenarios, we introduce multi-category anomalies into the images, as illustrated in Fig.~\ref{fig:dataset_distribution}.

\noindent\textbf{Phase 2: Depth Estimation and 3D Transformation.}
This phase combines image information with metric depth to transform 2D anomaly masks into precise 3D locations.

Specifically, we first utilize the large version of Depth Anything V2~\cite{yang2025depth} to compute pseudo-depth for both original and synthesized images. 
Subsequently, we randomly sample a predefined number of points from the point cloud and project them onto the image plane to obtain pixel coordinates $\mathbf{u}_i$ and Euclidean distances $d_i$ as:
\begin{equation}
\label{eq:projection}
\begin{split}
\mathbf{u}_i & = \frac{\mathbf{K} \cdot (\mathbf{R} \mathbf{p}_i + \mathbf{t})}{[\mathbf{R} \mathbf{p}_i + \mathbf{t}]_z}, \\
d_i &= \Vert \mathbf{R} \mathbf{p}_i + \mathbf{t} \Vert_{2},
\end{split}
\end{equation}
where $\mathbf{p}_i$ is a 3D point, $\mathbf{K}$ is the camera intrinsic matrix, $\mathbf{R}$ and $\mathbf{t}$ are the rotation and translation of the camera, $\mathbf{u}_i$ is the projected 2D point, and $d_i$ is the Euclidean distance.

By utilizing $ \mathbf{u}_i $ and the previously obtained depth map, we align the pseudo-depth with the real depth and apply Support Vector Regression (SVR) \cite{awad2015support} for nonlinear regression, mapping the pseudo-depth to real-world depth.
Finally, the aligned depth map and mask map from Phase 1 are used to derive 3D anomaly coordinates $(u, v, d)$, which are then transformed into the 3D voxel space.

\noindent\textbf{Phase 3: Voxelization and Scene Integration.}  
The pseudo-point clouds of anomalies are voxelized and integrated into the 3D voxel space of the original point cloud. 
Leveraging camera poses, we project each anomaly point $\mathbf{A}_i^v = (x_i^v, y_i^v, z_i^v)$ (representing the discrete voxel coordinates of the $i$-th anomaly) back into the scene to determine occlusion via ray marching. The ray position at step k is computed as:
\begin{equation}
\label{eq:ray_position}
\mathbf{c}_k = \mathbf{C}^v + k \cdot \frac{\mathbf{A}_i^v - \mathbf{C}^v}{\|\mathbf{A}_i^v - \mathbf{C}^v\|_2},
\end{equation}
where $k$ is the step index, $\mathbf{c}_k$ is the ray position at $k$, $\mathbf{C}^v$ is the camera position and $\mathbf{v}_k = \text{round}(\mathbf{c}_k)$ is the corresponding voxel center. The step size is controlled by a scaling factor $s > 0$, with $\Delta = \frac{1}{s}$ defining the resolution.

The occlusion process is formalized as:
\begin{equation}
\label{eq:voxel_projection}
\mathbf{V}[\mathbf{v}_k] =
\begin{cases}
1, & \text{if } \|\mathbf{c}_k - \mathbf{A}_i^v\|_2 \leq \Delta, \\
0, & \text{otherwise}.
\end{cases}
\end{equation}
where $0$ and $1$ represent the occupancy status of the voxel.

\subsection{VAA-KITTI, VAA-KITTI-360, and VAA-STU}

\begin{figure*}[!th]
    \centering
    \includegraphics[width=0.95\linewidth]{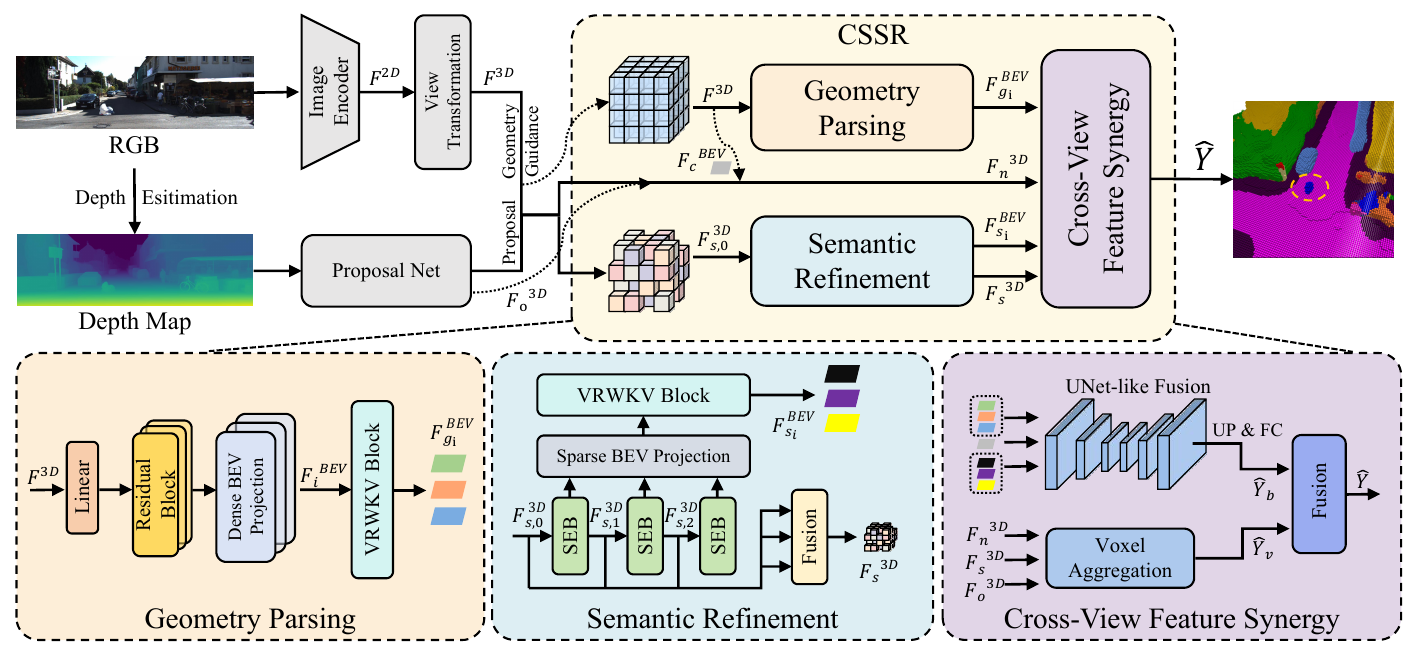}
    \vskip -1.5ex
    \caption{An overview of the proposed OccOoD framework. The image encoder extracts 2D features and converts them into 3D features via View Transformation, guided by geometric occupancy. Seed voxels and other features are processed through Geometry Parsing and Semantic Refinement to generate enhanced BEV and voxel features. Finally, Cross-View Feature Synergy integrates these representations for rich geometry and high spatial efficiency.}
    \vskip -3ex
    \label{fig:Overview of our framework}
\end{figure*}
We applied the aforementioned synthesis pipeline to SemanticKITTI~\cite{behley2019semantickitti} and SSCBench-KITTI-360~\cite{li2024sscbench}, creating two fresh datasets, VAA-KITTI and VAA-KITTI-360, which extend the original labels with 26 distinct anomaly categories. Being same-domain augmentations, they enable unsupervised OoD evaluation where performance gaps directly reflect detection capability rather than domain misalignment.

To ensure plausibility and naturalness, over $30,000$ images from the POC algorithm were carefully reviewed by three independent annotators.
After multiple rounds of cross-checking, images with poor generation quality, implausible object placements, or non-anomalous scenarios (\eg, \emph{birds flying in an urban sky or wheeled bins placed alongside sidewalks}) are excluded, yielding a final selection of $500$ images for both the VAA-KITTI and VAA-KITTI-360 test sets for the Out-of-Distribution Semantic Occupancy Prediction task.

In addition, we construct VAA-STU from STU~\cite{nekrasov2025stu}, a real-world 3D LiDAR dataset with annotated anomalous objects. Two sequences are voxelized into an occupancy-compatible format, preserving real-world complexity for robust OoD evaluation. Unlike the synthetic VAA-KITTI and VAA-KITTI-360, VAA-STU provides a real-world testbed for assessing generalization under natural OoD conditions.

\noindent\textbf{Regional-Level Metrics Design.}
Referring to 2D image anomaly detection~\cite{nayal2023rba}, we adopt Area under the Receiver Operating Characteristic Curve (AuROC) as a performance metric.
Furthermore, to mitigate the evaluation impact caused by the inaccuracy of existing methods in predicting the absolute positions of instances, and to bridge the gap between localization demands and anomaly-awareness goals, we propose a regional-level metric: Regional Area under the Precision-Recall Curve ($\text{AuPRC}_{r}$). The $\text{AuPRC}_{r}$ incorporates tolerance for minor localization inaccuracies while maintaining sensitivity to anomaly detection. 

\noindent\textbf{$\text{AuPRC}_{r}$:}  
The Precision-Recall curve evaluates the balance between precision and recall at varying thresholds:
\begin{equation}
\label{eq:10}
    \text{Precision}_r = \frac{\text{TP}_r}{\text{TP}_r + \text{FP}_r}, \quad
    \text{Recall}_r = \frac{\text{TP}_r}{\text{TP}_r + \text{FN}_r},
\end{equation}
where $\text{TP}_r$, $\text{FP}_r$, and $\text{FN}_r$ denote true positives, false positives, and false negatives within a regional tolerance radius $r$. The $\text{AuPRC}_{r}$ is computed as Eq.~(\ref{eq:11}):
\begin{equation}
\label{eq:11}
    \text{AuPRC}_r = \int_0^1 \text{Precision}_r(\text{Recall}_r) \, d(\text{Recall}_r).
\end{equation}

\section{OccOoD: Proposed Framework}

\subsection{Problem Setting}
Out-of-distribution segmentation in 2D pinhole images typically relies on pixel-wise anomaly scores.
Building on this concept, we extend out-of-distribution detection to 3D semantic occupancy prediction at the voxel level.
Given an input image, the semantic occupancy prediction process yields a volumetric representation \( V \in \mathbb{R}^{C \times H \times W \times Z} \), where $C$, $H$, $W$, and $Z$ correspond to the channels, height, width, and depth of the voxel space, respectively. The voxel-level classification scores \( Q(V) \in \mathbb{R}^{K \times H \times W \times Z} \) are then obtained, and $K$ denotes the number of classes for the semantic prediction. Ultimately, to assess anomalies at the voxel level, an anomaly score \( A(V) \in \mathbb{R}^{H \times W \times Z} \) is computed as Eq.~(\ref{eq:1}):
\begin{equation}
\label{eq:1}
A(V) = f(Q(V)).
\end{equation}

\subsection{Overall Architecture}

The OccOoD framework, illustrated in Fig.~\ref{fig:Overview of our framework}, integrates front-view and BEV representations for precise 3D semantic occupancy prediction.
3D voxels preserve fine geometric details, while BEV enhances spatial reasoning and global understanding. To leverage their complementary strengths, we propose the Cross-Space Semantic Refinement (CSSR), which extracts and refines features across both spaces. 
By fusing multi-scale representations across spaces, CSSR enhances feature representation, sharpens boundaries, and improves OoD detection.
Following SGN~\cite{mei2024camera}, 2D features extracted from RGB images are transformed into 3D voxel representations via a view transformation module based on MonoScene~\cite{cao2022monoscene}. These coarse features are refined through a geometry-guided auxiliary 3D occupancy head, supervised by a binary cross-entropy loss \( L_{\text{geo}} \), and subsequently filtered by a proposal network trained with an additional binary cross-entropy loss \( L_{\text{occ}} \).
The filtered features then undergo Geometry Parsing and Semantic Refinement to extract multi-scale BEV features encoding spatial structure and fine-grained semantic voxel features. These BEV features are fused with the original 3D features through a UNet-like module, ensuring precise alignment and enhanced learning. Finally, the upsampled BEV representation is combined with the aggregated voxel features through a 1D convolutional layer to produce the final semantic occupancy prediction.

\subsection{Cross-Space Semantic Refinement}
The coarse voxel features \( F^{3D} \) from the view transformation module and the seed features \( F_{\text{s,0}}^{3D} \) from the proposal network are in 3D voxel space. However, relying solely on voxel features limits semantic detail and boundary clarity, harming OoD detection. To address this, we propose Cross-Space Semantic Refinement (CSSR), which leverages voxel and BEV representations to enhance semantic prediction. Features are refined separately in the BEV space to capture global contextual structure and in the voxel space to preserve fine-grained local details and fused to produce accurate occupancy predictions with improved OoD sensitivity.

\noindent\textbf{Geometry Parsing.}
As shown in Fig.~\ref{fig:Overview of our framework}, the coarse voxel features \( F^{3D} \) are processed through a geometric parsing branch to extract compact BEV representations encoding spatial structure. This branch projects \( F^{3D} \) into a lower-dimensional space, applies residual refinement and z-axis alignment, then compresses it via 2D convolutions to generate dense BEV features. These are enhanced by the Vision-RWKV (VRWKV) module~\cite{duan2024vision} through spatial and channel mixing, yielding dense geometric BEV features \( \{ F_{\text{g}_i}^{\text{BEV}} \}_i \).

\noindent\textbf{Semantic Refinement.} 
The seed voxels \(F_{\text{s,0}}^{3D}\) from the proposal network undergo semantic refinement to enhance accuracy.
The module comprises three cascaded sparse encoder blocks (SEBs), each linked with a BEV projection and a VRWKV block. 
Each SEB contains a Sparse Feature Encoder (SFE)~\cite{ye2022efficient} and a Sparse Geometry Feature Encoder (SGFE)~\cite{ye2022efficient}, which refine the initial representation through hierarchical pooling and attention, capturing fine-grained details and contextual dependencies.

First, the multi-scale features \(F_{\text{s,1}}^{3D}\) and \(F_{\text{s,2}}^{3D}\) output by the first two SEBs are projected into BEV space via max-pooling, generating sparse features. These are densified using Spconv and enhanced by the VRWKV module~\cite{duan2024vision} to model long-range dependencies, yielding multi-scale semantic BEV features \( \{ F_{\text{s}_i}^{\text{BEV}} \}_i \).
The refined voxel features \(F_{\text{s}}^{3D}\) are obtained by fusing \(F_{\text{s,0}}^{3D}\), \(F_{\text{s,1}}^{3D}\), and \(F_{\text{s,2}}^{3D}\) via an MLP (Eq.~\ref{eq:f3ds}):
\begin{equation}
\label{eq:f3ds}
F_{\text{s}}^{3D}= \text{MLP}([F_{\text{s,0}}^{3D}, F_{\text{s,1}}^{3D}, F_{\text{s,2}}^{3D}]).
\end{equation}

The multi-scale semantic BEV features \( \{ F_{\text{s}_i}^{\text{BEV}} \}_i \) are used for BEV-level fusion, while the refined semantic feature \(F_{\text{s}}^{3D}\) is utilized for voxel-level fusion. Additionally, \(F_{\text{s}}^{3D}\) is passed through an auxiliary semantic head (a two-layer MLP) to predict the corresponding semantics \(\hat{Y}_s\).

The semantic refinement stage is optimized using a combined loss of cross-entropy and Lovasz loss~\cite{berman2018lovasz} as follows:
\begin{equation}
\label{eq:5}
L_{sem} = L_{ce}(\hat{Y}_s, Y_s) + L_{lovasz}(\hat{Y}_s, Y_s),
\end{equation}
where \(Y_s\) represents the semantic labels of seed voxels. 

\begin{table*}[!t]
    \centering
    \caption{Quantitative results on the SemanticKITTI validation set~\cite{behley2019semantickitti} and VAA-KITTI. The best and second-best results are shown in \textbf{bold} and \underline{under lined}, respectively.}
    \vskip -1ex
    \resizebox{0.90\textwidth}{!}{
        \fontsize{7.45pt}{11pt}\selectfont
        \setlength{\tabcolsep}{1.2pt} 
        \begin{tabularx}{0.84\textwidth}{@{}l|c|*{19}{c}|c@{}}
            \toprule
            Method 
            &IoU
            & \rotatebox{90}{{\textcolor{red}{\rule{0.5em}{0.5em}}}\hspace{0.3em}road} 
            & \rotatebox{90}{{\textcolor{pink}{\rule{0.5em}{0.5em}}}\hspace{0.3em}sidewalk} 
            & \rotatebox{90}{{\textcolor{orange}{\rule{0.5em}{0.5em}}}\hspace{0.3em}parking} 
            & \rotatebox{90}{{\textcolor{yellow}{\rule{0.5em}{0.5em}}}\hspace{0.3em}otherground} 
            & \rotatebox{90}{{\textcolor{green}{\rule{0.5em}{0.5em}}}\hspace{0.3em}building} 
            & \rotatebox{90}{{\textcolor{blue}{\rule{0.5em}{0.5em}}}\hspace{0.3em}car} 
            & \rotatebox{90}{{\textcolor{purple}{\rule{0.5em}{0.5em}}}\hspace{0.3em}truck} 
            & \rotatebox{90}{{\textcolor{brown}{\rule{0.5em}{0.5em}}}\hspace{0.3em}bicycle} 
            & \rotatebox{90}{{\textcolor{cyan}{\rule{0.5em}{0.5em}}}\hspace{0.3em}motorcycle} 
            & \rotatebox{90}{{\textcolor{magenta}{\rule{0.5em}{0.5em}}}\hspace{0.3em}othervehicle} 
            & \rotatebox{90}{{\textcolor{lime}{\rule{0.5em}{0.5em}}}\hspace{0.3em}vegetation} 
            & \rotatebox{90}{{\textcolor{teal}{\rule{0.5em}{0.5em}}}\hspace{0.3em}trunk} 
            & \rotatebox{90}{{\textcolor{violet}{\rule{0.5em}{0.5em}}}\hspace{0.3em}terrain} 
            & \rotatebox{90}{{\textcolor{gray}{\rule{0.5em}{0.5em}}}\hspace{0.3em}person} 
            & \rotatebox{90}{{\textcolor{black}{\rule{0.5em}{0.5em}}}\hspace{0.3em}bicyclist} 
            & \rotatebox{90}{{\textcolor{lightgray}{\rule{0.5em}{0.5em}}}\hspace{0.3em}motorcyclist} 
            & \rotatebox{90}{{\textcolor{olive}{\rule{0.5em}{0.5em}}}\hspace{0.3em}fence} 
            & \rotatebox{90}{{\textcolor{navy}{\rule{0.5em}{0.5em}}}\hspace{0.3em}pole} 
            & \rotatebox{90}{{\textcolor{maroon}{\rule{0.5em}{0.5em}}}\hspace{0.3em}trafficsign} 
            & mIoU \\
            \midrule
            \multicolumn{22}{c}{SemanticKITTI}\\
            \midrule       
            MonoScene~\cite{cao2022monoscene}

            & 37.12 & 57.47 & 27.05 & 15.72 
            & \textbf{0.87} & 14.24 & 23.55 & 7.83 & 0.20 & 0.77 & 3.59
            & 18.12 & 2.57 & 30.76 & \underline{1.79} & \underline{1.03} & 0.00 
            & 6.39 & 4.11 & 2.48 & 11.50 \\

            SGN~\cite{mei2024camera}  
            & \underline{46.21} & 59.10 & 29.41 & 19.05 
            & \underline{0.33} & \underline{25.17} & 33.31 & 6.03 & 0.61 & 0.46 & \underline{9.84}
            & \textbf{28.93} & \textbf{9.58} & \underline{38.12} & 0.47 & 0.10 & 0.00 
            & \textbf{9.96} & \textbf{13.25} & \underline{7.21} & 15.32 \\

            CGFormer~\cite{yu2024context}
            & 45.99 &\textbf{65.51} &\textbf{32.31} &\underline{20.82} &0.16 &23.52 &\textbf{34.32} &\underline{19.44} &\textbf{4.61} &\textbf{2.71} &7.67 &26.93 &8.83 &\textbf{39.54} &\textbf{2.38} &\textbf{4.08} &0.00 &9.20 &10.67 &\textbf{7.84}&\textbf{16.89}\\
            \midrule
            OccOoD (Ours) 
            & \textbf{46.83} & \underline{60.87} & \underline{32.16} & \textbf{23.52} 
            &  0.06 &  \textbf{26.54} & \underline{33.51} & \textbf{25.17} & \underline{0.65} & \underline{1.00} & \textbf{11.84} 
            & \underline{28.43} & \underline{8.90} & 36.96 & 0.44 & 0.16 & 0.00 
            & \underline{9.94} & \underline{13.00} & 6.03 & \underline{16.80} \\
            \midrule
            \multicolumn{22}{c}{VAA-KITTI}\\
            \midrule         
            SGN~\cite{mei2024camera}&\underline{39.07}  &\underline{46.62} &\underline{26.54} &\underline{15.88} &\textbf{0.12} &\textbf{18.34} &\textbf{20.51} &\underline{0.62} &\textbf{0.44} &\underline{0.35} &\textbf{1.23} &\underline{24.76} &\textbf{6.40} &\underline{21.78}&\underline{0.48} &\underline{1.29} &0.00 &\textbf{12.25} &\textbf{8.70} &\underline{4.57}&\underline{11.10}
            \\            
            OccOoD (Ours)&\textbf{40.04}  &\textbf{50.17} &\textbf{27.42} &\textbf{18.40} &\underline{0.01} &\underline{17.38} &\underline{20.19} &\textbf{0.74} &\underline{0.26} &\textbf{0.56} &\underline{1.04} &\textbf{25.13} &\underline{5.53} &\textbf{23.56}&\textbf{0.67} &\textbf{1.95} &0.00 &\underline{11.35} &\underline{8.15} &\textbf{5.80}&\textbf{11.49}
            \\           
            \bottomrule
        \end{tabularx}
        }
        \vskip -2ex
    \label{table:kitti_val}
\end{table*}

\begin{table}[t]
    \centering
    \caption{OoD detection performance on VAA-KITTI dataset. Best results in \textbf{bold}.}
    \begin{adjustbox}{width=0.4\textwidth, center}
    \begin{tabular}{@{}l|ccc|c@{}}
    \toprule
    &\multicolumn{4}{c}{VAA-KITTI}\\
    \multirow{2}{*}{Methods} & \multicolumn{3}{c|}{AuPRC$_r$$\uparrow$} & \multirow{2}{*}{AuROC$\uparrow$}\\
     & 0.8m & 1.0m & 1.2m & \\
    \midrule
    MonoScene~\cite{cao2022monoscene} & 2.81 & 3.48 & 4.31 & 63.54  \\
    CGFormer~\cite{yu2024context} & 5.65 & 6.44 &7.36  &  64.09 \\   
    SGN~\cite{mei2024camera} & 12.69 & 18.81 & 27.10 & 65.40  \\
    \midrule
    OccOoD (ours) & \textbf{17.15} & \textbf{23.48} & \textbf{31.83}  & \textbf{65.50} \\     
    \bottomrule
    \end{tabular}
    \end{adjustbox}
    \vskip -4ex
    \label{tab:vaa-kitti}
\end{table}

\noindent\textbf{Cross-View Feature Synergy.}  
To integrate BEV and voxel-level representations, we process them in parallel.

For the BEV branch, a UNet-like architecture encodes hierarchical spatial features and decodes high-resolution maps for fusion, including the projected 3D features \(F^\mathrm{BEV}_c\), geometric BEV features \( \{ F_{\text{g}_i}^{\text{BEV}} \}_i \), and semantic BEV features \( \{ F_{\text{s}_i}^{\text{BEV}} \}_i \).
The encoder downsamples \( F^\mathrm{BEV}_c \) through convolutional blocks, each followed by a VRWKV block, and fuses the three BEV streams using an ARF module~\cite{mei2023ssc}: 
\begin{equation}
\label{eq:Y_b}
F^{\text{BEV}}_{f_i} = \smashoperator{\sum_{k \in \{\text{g}, \text{s}, \text{c}\}}} \Big( F^{\text{BEV}}_{k_i} \odot A^{\text{BEV}}_{k_i} \Big),
\end{equation}
where \(F^{\text{BEV}}_{k_i}\) and \(A^{\text{BEV}}_{k_i}\) denote the $k$-th feature map and its attention weight, respectively. The decoder restores resolution via transposed convolutions and skip connections, and the output is reshaped and upsampled to yield a voxel-level representation $\hat{Y}_{b}$.

For the voxel branch, we follow SGN~\cite{mei2024camera}, fusing refined semantic features \(F_{\text{s}}^{3D}\), non-seed features \(F_{\text{n}}^{3D}\), and occupancy features \(F^{3D}_o\) via Multi-Scale Semantic Propagation (MSSP) as follows:
\begin{equation}
\label{eq:ff}
\hat{Y}_{\text{$v$}}= \text{MSSP}([F_{\text{s}}^{3D}, F_{\text{n}}^{3D}, F_{\text{$o$}}^{3D}]).
\end{equation}

Subsequently, $\hat{Y}_{v}$ and $\hat{Y}_{b}$ are fused via a 1D convolutional layer to produce the final prediction $\hat{Y}_f$. The model is optimized with a combined loss as follows:
\begin{equation}
\label{eq:6}
L_{ssc} = L_{sem}^{scal}(\hat{Y}, Y) + L_{geo}^{scal}(\hat{Y}, Y) + L_{ce}(\hat{Y}, Y),
\end{equation}
where $L_{sem}^{scal}$ ensures semantic consistency at both scene and class levels, $L_{geo}^{scal}$ enforces geometric fidelity, and $L_{ce}$ minimizes classification errors.

\noindent\textbf{Loss Function.} 
The total loss is defined as:
\begin{equation}
\label{eq:7}
L = L_{geo} + L_{occ} + L_{sem} + L_{ssc},
\end{equation}
where equal weights are applied to balance all tasks.

\subsection{OoD Occupancy Prediction}
After achieving high-precision semantic occupancy prediction, we further integrate an out-of-distribution detection mechanism to identify outliers that deviate from the in-distribution regions. For the semantic occupancy prediction results $\hat{Y}_f$, we compute the category-specific anomaly score \( A(V) \) for each voxel according to the predictive semantics.

\noindent\textbf{Geometry Prior.}
 To more efficiently focus on non-empty regions, we introduce a geometry prior by applying an \emph{argmax} operation to $\hat{Y}_f$.
 This strategy prioritizes potentially anomalous non-empty regions, significantly improving the effectiveness and accuracy of OoD detection.

\noindent\textbf{Anomaly-Sensitive Scoring.}
We compute anomaly scores based on the predicted semantic category, employing distinct strategies to capture the two primary forms of uncertainty pertinent to OoD detection.
\begin{itemize}
    \item For \textbf{instance} (\eg, \emph{car, person}), which form compact clusters in the logit space, anomalies appear as outliers from the class centroid. We measure this deviation using \emph{cosine similarity}:
    \begin{equation}
        \label{eq:cosine}
        S_{\text{cosine}}(v_i) = 1 - \cos(\mathbf{l}(v_i), \bar{\mathbf{l}}),
    \end{equation}
    where $\mathbf{l}(v_i)$ is the logit vector of voxel $v_i$, and $\bar{\mathbf{l}}$ is the mean logit vector of the same class. 
    \item For \textbf{region} (\eg, \emph{road, vegetation}), which occupy large, homogeneous areas, OoD regions typically induce high prediction uncertainty. This is quantified using \emph{entropy}:
    \begin{equation}
        \label{eq:entropy}
        S_{\text{entropy}}(v_i) = - \sum_{k \in \mathcal{K}} p(k | v_i) \log p(k | v_i),
    \end{equation}
    where $\mathcal{K}$ represents the set of possible classes, and $p(k | v_i)$ is the predicted probability of class $k$ for voxel $v_i$.
To prioritize detection around structured instance classes, the entropy-based scores for region classes are down-weighted. Higher entropy indicates greater uncertainty, a hallmark of OoD regions.
\end{itemize}

\begin{table}[t]
    \centering
    \caption{OoD detection performance on VAA-KITTI-360 and VAA-STU datasets. Best results in \textbf{bold}.}
    \resizebox{0.475\textwidth}{!}{%
    \begin{tabular}{@{}l|ccc|c|ccc|c@{}}
    \toprule
    &\multicolumn{4}{c|}{VAA-KITTI-360}&\multicolumn{4}{c}{VAA-STU}\\
    \multirow{2}{*}{Methods} & \multicolumn{3}{c|}{AuPRC$_r$$\uparrow$} & \multirow{2}{*}{AuROC$\uparrow$}& \multicolumn{3}{c|}{AuPRC$_r$$\uparrow$} & \multirow{2}{*}{AuROC$\uparrow$}\\
     & 0.8m & 1.0m & 1.2m & & 0.8m & 1.0m & 1.2m&\\
    \midrule
    SGN~\cite{mei2024camera} & 12.24 & 24.89 & 45.15 & 45.06 & 2.91 & 5.74  & 11.32 & 64.51  \\
    \midrule
    OccOoD (ours) & \textbf{12.84} & \textbf{25.74} & \textbf{46.28}  & \textbf{47.90} & \textbf{3.05} & \textbf{6.15} & \textbf{11.39}  & \textbf{65.37}\\     
    \bottomrule
    \end{tabular}}
    \vskip -4ex
    \label{tab:vaa-kitti-360-stu}
\end{table}

Ultimately, the anomaly scores $A(V)$ are normalized to $[0, 1]$ to ensure comparability across classes. Voxels predicted as empty are assigned the minimum anomaly score, effectively filtering them from further analysis.

\section{Experiments}
\subsection{Implementation Details}
We train OccOoD ($38M$ parameters) for $35$ epochs on four RTX3090 GPUs with a batch size of $4$, using AdamW with a learning rate of $2e{-}4$ and weight decay of $1e{-}2$. We evaluate OccOoD’s semantic occupancy prediction on public benchmarks. For OoD detection, our evaluation is conducted on the proposed datasets with single‑frame settings.

\subsection{Quantitative Results}

\label{sec:ood_results}
\noindent\textbf{Out-of-Distribution Detection.} In Tables \ref{tab:vaa-kitti} and \ref{tab:vaa-kitti-360-stu}, the OoD detection results are reported. Across the three scales of AuPRC$_r$, our network achieves a significant improvement over the baseline. On the VAA-KITTI dataset, our performance improves by more than $4\%$ in terms of AuPRC$_r$. Similarly, on both the VAA-KITTI-360 and VAA-STU datasets, our method demonstrates superior performance. Notably, CGFormer~\cite{yu2024context} underperforms in OoD detection despite strong in-distribution accuracy. When encountering anomalies (\eg, \emph{animals}, \emph{garbage bags}), it often assigns them to rare in-distribution categories such as \emph{bicycle} and \emph{motorcycle} with high confidence, lacking clear semantic correspondence. This overconfident misclassification reflects poor model calibration and an inability to recognize distributional novelty. By enforcing consistency between voxel and BEV spaces, CSSR reduces overconfident misclassification and enhances sensitivity to anomalies.

\begin{table*}[!t]
    \centering
    \caption{Quantitative results on the SSCBench-KITTI-360 test set~\cite{li2024sscbench} and VAA-KITTI-360. Results for the compared methods are taken from~\cite{li2024sscbench}. The best and second-best results are in \textbf{bold} and \underline{under lined}, respectively.}
    \vspace{-1ex}

    \resizebox{0.9\textwidth}{!}{
        \fontsize{7.8pt}{10pt}\selectfont
        \setlength{\tabcolsep}{1.2pt} 
        \begin{tabularx}{0.84\textwidth}{@{}l|c|*{18}{c}|c@{}}
            \toprule
            Method 
            &IoU
            & \rotatebox{90}{{\textcolor{blue}{\rule{0.5em}{0.5em}}}\hspace{0.3em}car} 
            & \rotatebox{90}{{\textcolor{brown}{\rule{0.5em}{0.5em}}}\hspace{0.3em}bicycle} 
            & \rotatebox{90}{{\textcolor{cyan}{\rule{0.5em}{0.5em}}}\hspace{0.3em}motorcycle} 
            & \rotatebox{90}{{\textcolor{purple}{\rule{0.5em}{0.5em}}}\hspace{0.3em}truck} 
            & \rotatebox{90}{{\textcolor{magenta}{\rule{0.5em}{0.5em}}}\hspace{0.3em}other-veh.} 
            & \rotatebox{90}{{\textcolor{gray}{\rule{0.5em}{0.5em}}}\hspace{0.3em}person} 
            & \rotatebox{90}{{\textcolor{red}{\rule{0.5em}{0.5em}}}\hspace{0.3em}road} 
            & \rotatebox{90}{{\textcolor{orange}{\rule{0.5em}{0.5em}}}\hspace{0.3em}parking}           
            & \rotatebox{90}{{\textcolor{pink}{\rule{0.5em}{0.5em}}}\hspace{0.3em}sidewalk} 
            & \rotatebox{90}{{\textcolor{yellow}{\rule{0.5em}{0.5em}}}\hspace{0.3em}other-grnd.} 
            & \rotatebox{90}{{\textcolor{green}{\rule{0.5em}{0.5em}}}\hspace{0.3em}building} 
            & \rotatebox{90}{{\textcolor{olive}{\rule{0.5em}{0.5em}}}\hspace{0.3em}fence} 
            & \rotatebox{90}{{\textcolor{lime}{\rule{0.5em}{0.5em}}}\hspace{0.3em}vegetation} 
            & \rotatebox{90}{{\textcolor{violet}{\rule{0.5em}{0.5em}}}\hspace{0.3em}terrain} 
            & \rotatebox{90}{{\textcolor{navy}{\rule{0.5em}{0.5em}}}\hspace{0.3em}pole} 
            & \rotatebox{90}{{\textcolor{maroon}{\rule{0.5em}{0.5em}}}\hspace{0.3em}traf.sign} 
            & \rotatebox{90}{{\textcolor{blue}{\rule{0.5em}{0.5em}}}\hspace{0.3em}other-struct.}             
            & \rotatebox{90}{{\textcolor{teal}{\rule{0.5em}{0.5em}}}\hspace{0.3em}other-obj.} 

            & mIoU \\
            \midrule
            \multicolumn{21}{c}{SSCBench-KITTI-360}\\
            \midrule

             MonoScene~\cite{cao2022monoscene} &37.87  &19.34 &0.43 &0.58 &8.02 &2.03 &0.86 &48.35 &11.38 &28.13 &3.32 &32.89 &3.53 &26.15 &16.75 &6.92 &5.67 &4.20 &3.09 &12.31 \\
            
            VoxFormer~\cite{li2023voxformer} & 38.76  & 17.84 & 1.16 & 0.89 & 4.56 & 2.06 & 1.63 & 47.01 & 9.67 & 27.21 & 2.89 & 31.18 & 4.97 & 28.99 & 14.69 & 6.51 & 6.92 & 3.79 & 2.43  & 11.91\\
            
            SGN~\cite{mei2024camera} &\underline{47.06}  &\underline{29.03} &\textbf{3.43}&\underline{2.90} &\underline{10.89} &\underline{5.20} &2.99&58.14&15.04&36.40&4.43 &42.02&7.72&\underline{38.17}&\underline{23.22} &\underline{16.73} &16.38 &\underline{9.93} &5.86&18.25\\       
            CGFormer~\cite{yu2024context} &\textbf{48.07}  &\underline{29.85} &\underline{3.42} &\textbf{3.96} &\underline{17.59} &\textbf{6.79} &\textbf{6.63} &\textbf{63.85} &\textbf{17.15} &\textbf{40.72} &\textbf{5.53} &\underline{42.73} &\textbf{8.22} &\textbf{38.80} &\textbf{24.94} &16.24 &\underline{17.45} &\textbf{10.18} &\underline{6.77}&\textbf{20.05}\\

            \midrule
            OccOoD~(Ours)
            &46.93 &28.51 &2.95 &0.94 &9.48 &4.57 &\underline{2.99} &\underline{58.91} &\underline{16.17} &\underline{37.04} &\underline5.34 &\textbf{42.99} &\underline{7.83} &37.41 &23.01 &\textbf{17.47} &\textbf{18.87} &8.94 &\textbf{7.16} &\underline{18.37}\\
            \midrule
            \multicolumn{21}{c}{VAA-KITTI-360}\\
            \midrule         
            SGN~\cite{mei2024camera} &\textbf{39.90} &\underline{21.93} &\textbf{0.04} &0.00 &0.00 &\underline{0.66} &\textbf{0.88} &\underline{46.95} &\underline{13.81} &\underline{25.67} &\underline{2.72} &\textbf{33.31} &\textbf{3.07} &\textbf{28.25}&\textbf{12.63} &\textbf{8.95} &\textbf{13.84} &\textbf{3.79}&\underline{1.52} &\underline{12.11}
            \\            
            OccOoD (Ours)&\underline{39.26} &\textbf{22.84} &\underline{0.00} &0.00 &0.00 &\textbf{3.54} &\underline{0.58} &\textbf{49.09} &\textbf{14.65} &\textbf{27.84} &\textbf{3.41} &\underline{32.59} &\underline{1.55} &\underline{27.46}&\underline{11.46} &\underline{8.42} &\underline{12.30}  &\underline{3.63}&\textbf{6.49} &\textbf{12.55}
            \\            
            \bottomrule
        \end{tabularx}
        }

        \label{table:kitti-360}
        \vskip -2ex
\end{table*}

\begin{figure*}[!t]
    \centering
    \includegraphics[width=0.75\linewidth]{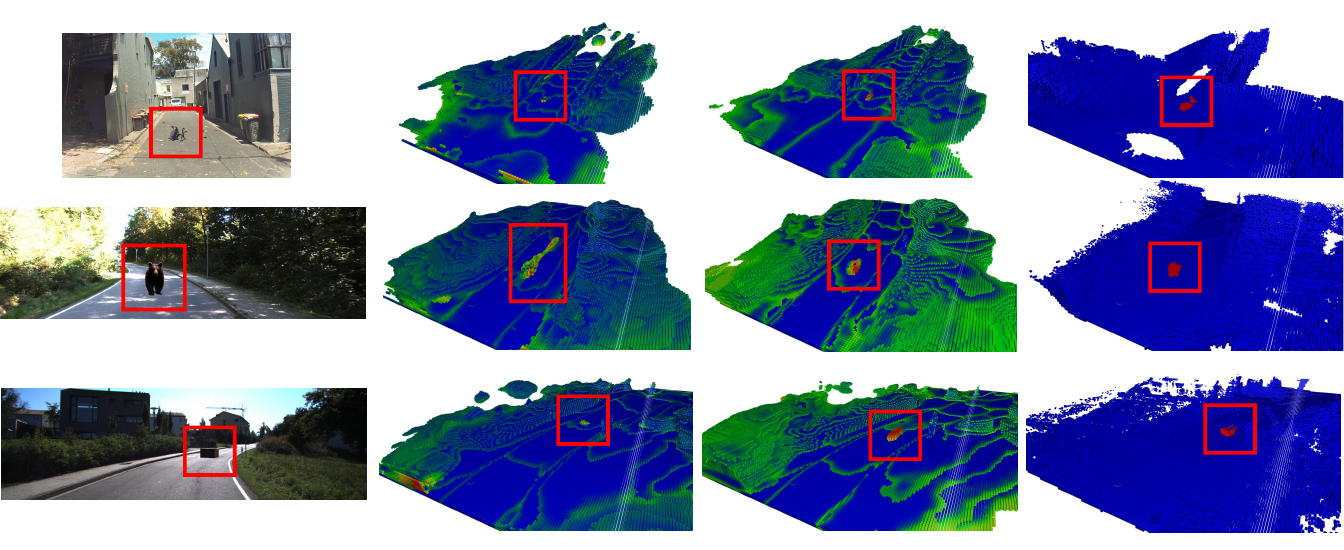}
    \vskip -2ex
    \caption{Visualization of the results on VAA-KITTI and VAA-STU datasets. From left to right are the input RGB images, the SGN~\cite{mei2024camera} results, the OccOoD (ours) results, and the OoD ground truths. \textbf{Zoom in for a better view.}}
    \label{fig:visual}
    \vskip -3ex
\end{figure*}

\noindent\textbf{3D Occupancy Prediction.} We present the experimental results of semantic occupancy prediction in Table \ref{table:kitti_val} and Table \ref{table:kitti-360}. Our method consistently demonstrates performance improvements over the baseline model, regardless of whether the evaluation is conducted on public datasets or our self-constructed datasets. These results validate the effectiveness of our designed CSSR, which leverages complementary strengths of voxel and BEV representations to enhance semantic refinement and global context modeling.

\subsection{Ablation Studies}
\label{sec:ablation_studies}

\begin{table}[t]
    \centering

    \caption{Ablation studies of ASS on VAA-KITTI.}
    \vskip -1ex
    \begin{tabular}{@{}c|ccc|c}
        \toprule
        \multirow{3}{*}{\makecell{Anomaly Score \\ Methods}} & \multicolumn{4}{c}{VAA-KITTI}\\
        & \multicolumn{3}{c|}{AuPRC$_r$$\uparrow$} & \multirow{2}{*}{AuROC$\uparrow$} \\
         & 0.8m & 1.0m & 1.2m &\\
        \midrule
        Entropy & 7.11 & 12.26 & 19.95 & 65.19 \\
        Postpro & 5.83 & 9.92 & 15.94 & 64.88 \\
        Energy & 7.24 & 12.40 & 20.05 & 64.78 \\
        \midrule
        ASS (Ours)& \textbf{17.15} & \textbf{23.48} & \textbf{31.83} & \textbf{65.50} \\
        \bottomrule
    \end{tabular}
    \label{tab:ood_vaa-kitti_modified}
    \vskip -6ex
\end{table}
\noindent\textbf{Anomaly-Sensitive Scoring.} 
Table~\ref{tab:ood_vaa-kitti_modified} demonstrates the effectiveness of our proposed Anomaly-Sensitive Scoring (ASS). Compared to the other three scoring methods, our approach improves AuPRC$_r$ by more than $10\%$ across all scales. This improvement stems from our dual-strategy design, which applies cosine similarity to compact instances and entropy to homogeneous regions, effectively matching the scoring to each object's spatial characteristics for better OoD identification.

\noindent\textbf{Geometry Prior.} 
Table~\ref{tab:ood_vaa-kitti_modified_geometry_prior} evaluates the role of the geometry prior via AuPRC$_r$ and AuROC. Including the prior substantially improves AuPRC$_r$. Without it, the higher AuROC arises because empty voxel uncertainty is also scored as anomalous. Given the scarcity of positive samples in our dataset, AuPRC$_r$ is the more appropriate performance metric.

\subsection{Qualitative Results}
\noindent\textbf{Qualitative Comparison.} 
As shown in Fig.~\ref{fig:visual}, unlike SGN~\cite{mei2024camera} which solely relies on voxel input and produces inconsistent shape estimations, our method leverages the complementary strengths of voxel and BEV features through CSSR, effectively improving the localization of small objects (\eg, \emph{chair}) and maintaining the structural continuity of larger ones (\eg, \emph{bear}), yielding more complete and geometrically coherent occupancy estimates and thereby improving OoD detection performance.

\begin{table}[t]
    \centering
    \caption{Ablation studies of geometry prior within $1.2 m$.}
    \vskip -1ex
    \begin{adjustbox}{width=0.95\linewidth}
    \begin{tabular}{@{}c|c|c|c|c@{}}
        \toprule
        &\multicolumn{2}{c|}{VAA-KITTI}&\multicolumn{2}{c}{VAA-KITTI-360}\\
        
        \multirow{1}{*}{Methods} &AuPRC$_r$$\uparrow$ & \multirow{1}{*}{AuROC$\uparrow$}&AuPRC$_r$$\uparrow$ & \multirow{1}{*}{AuROC$\uparrow$} \\
        
        \midrule
        $w/o$ Geometry prior & 14.86 & \textbf{76.59} & 35.17 & 36.72 \\
        \midrule
        Geometry prior& \textbf{31.83} & 65.50 & \textbf{46.28} & \textbf{47.90}\\
        \bottomrule
    \end{tabular}
    \end{adjustbox}

    \label{tab:ood_vaa-kitti_modified_geometry_prior}
    \vskip -2ex
\end{table}

\begin{figure}[!t]
    \centering
    \includegraphics[width=0.95\linewidth]{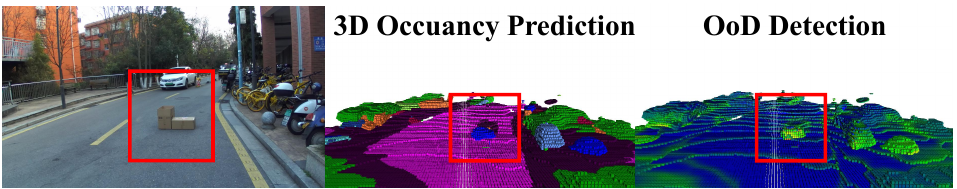}
    \vskip -1ex
    \caption{Results in real-world urban environments.}
    \label{fig:real}
    \vskip -5ex
\end{figure}

\noindent\textbf{Insights Into Real-World Applications.} 
Fig.~\ref{fig:real} presents results from additional real-world urban scenes not included in VAA-STU. We collected these real urban scenes using a ZED Mini camera and ran inference on the left-view images. The model demonstrates practical efficiency, achieving $9$ FPS on a GPU RTX3090. The left image contains an OoD object (a box on the road). The 3D occupancy prediction (middle) misclassifies it as a motorcycle, exposing a key limitation. In contrast, our OoD detector (right) correctly identifies this anomaly with high confidence, demonstrating its potential to enhance autonomous system safety.
\section{Conclusion}
In this paper, we look into out-of-distribution semantic occupancy prediction to address OoD voxel detection in 3D space.
To address the lack of anomaly annotations in existing datasets, we propose a Realistic Anomaly Augmentation and develop three evaluation datasets: VAA-KITTI, VAA-KITTI-360, and VAA-STU. 
Our method, OccOoD, integrates OoD detection into 3D semantic occupancy prediction by combining voxel-based and BEV representations, effectively capturing both local detail and global context for enhanced semantic and geometric reasoning. Experiments show that OccOoD achieves superior OoD detection performance on these datasets while maintaining competitive occupancy prediction accuracy on standard benchmarks.

\bibliographystyle{IEEEtran}
\bibliography{IEEEfull}

\end{document}